\documentclass[11pt,draftcls,onecolumn,journal]{IEEEtran}
\usepackage{amsmath,amsfonts}
\usepackage{algorithm}
\usepackage{array}
\usepackage[caption=false,font=normalsize,labelfont=sf,textfont=sf]{subfig}
\usepackage{textcomp}
\usepackage{stfloats}
\usepackage{url}
\usepackage{verbatim}
\usepackage{graphicx}
\usepackage{cite}
\usepackage{hyperref}
\usepackage{booktabs}
\usepackage{geometry}      % For page margins
\usepackage{lscape}        % For the landscape environment
\usepackage{multirow}      % For the \multirow command
\usepackage{amssymb}       % For the \checkmark symbol
\usepackage{algpseudocode}
\usepackage{threeparttable}
\usepackage{orcidlink}

\usepackage{tikz}
\usepackage{xcolor}

\definecolor{patchred}{RGB}{255,204,204}
\definecolor{patchblue}{RGB}{218,232,252}

\DeclareRobustCommand{\basesquare}{%
    \tikz[baseline=-0.6ex]{\draw[fill=patchred, draw=red!60!black, line width=0.6pt] (-0.1,-0.1) rectangle (0.1,0.1);}%
}
\DeclareRobustCommand{\targetcircle}{%
    \tikz[baseline=-0.6ex]{\draw[fill=patchblue, draw=blue!60!black, line width=0.6pt] (0,0) circle (0.12);}%
}

\usepackage[normalem]{ulem}

\algrenewcommand{\algorithmiccomment}[1]{\hfill // \texttt{#1}}

\begin{document}

\title{Mechanistic Interpretability of Deep Neural Networks: Methods, Frameworks, and Perspectives}

\author{Lorenzo Cirillo\orcidlink{0009-0008-5266-8957}, Irene Amerini\orcidlink{0000-0002-6461-1391}, Andrea Cavallaro\orcidlink{0000-0001-5086-7858}}% <-this % stops a space
% }



\maketitle

{\renewcommand{\thefootnote}{}%
 \footnotetext{\copyright~2026 IEEE. Personal use of this material is permitted.
 Permission from IEEE must be obtained for all other uses, in any current or future
 media, including reprinting/republishing this material for advertising or promotional
 purposes, creating new collective works, for resale or redistribution to servers or
 lists, or reuse of any copyrighted component of this work in other works.}}

\section*{}
\label{sec:abstract}

Deep neural networks underpin fundamental tasks in a wide range of signal domains, such as image, audio, and language processing. However, their internal mechanisms are notably difficult to interpret.  Internal activations are structured signals transformed across layers that may encode a multitude of interdependent relationships, as individual neurons frequently represent multiple underlying concepts. 
Mechanistic interpretability aims to uncover the internal mechanisms of neural networks by identifying components whose activations are causally responsible for specific behaviors. In such a context, this article provides an overview of advances in mechanistic interpretability for the analysis of causal inference in deep neural networks. 

%%%%%%%%%%%%%%%%%%%%%%%%%%%%%%%%%%%%%%%%%%%%%%%%%%%%%%%%%%%%%%%%%%%%%%%%%%%

\section*{Introduction}
\label{sec:introduction}

\IEEEPARstart{D}eep neural networks have achieved unprecedented performance across several signal domains, including image, audio, and language processing~\cite{spm1, dm_magazine}. These networks perform complex tasks such as classification,  detection, generation, and forecasting. However, their internal processing mechanisms are difficult to interpret~\cite{conmy2023towards}. 
Uncovering causal relationships in deep models is fundamental for improving reliability, controllability, safety, alignment, and mitigating the risk of undesirable behaviors of learning systems~\cite{ji}. Building on this, traditional interpretability and mechanistic interpretability aim to make the decision-making process of neural networks understandable, and their characteristics are compared in Table~\ref{tab:xai_vs_mi}.

Traditional interpretability techniques focus on input-output relationships and high-level summaries, operate on the relevance of the input features to the predicted final output, and identify which input features most influence the model output. Notable approaches include integrated gradients~\cite{integrated_gradients} and GradCAM~\cite{gradcam}.
Interpretability techniques can be model-agnostic, such as integrated gradients, which apply to any network regardless of its components, or model-specific, such as GradCAM, which is tailored for convolutional neural networks. Although these methods consider the gradients of the network to attribute input feature importance, they do not explicitly examine the activations of internal model components, offering limited insights into how latent representations are constructed through the model architecture.

In contrast, mechanistic interpretability aims to identify the internal mechanisms of neural networks, relying on model components responsible for specific behaviors~\cite{mi_survey_4}. A model component is an entity within the neural network that takes a signal as input, processes it, and outputs a corresponding activation. Examples of model components include neurons, layers, and modules. The components analyzed by a mechanistic interpretability method define its granularity, which refers to the resolution at which internal mechanisms are identified. At the finest scale, mechanistic methods target the causal contribution of specific neurons. At coarser scales, the focus shifts to aggregated transformations applied by entire architectural modules or layers to predict the model output.
An internal mechanism refers to the causal relationship between model
components and inputs for a given model behavior. For instance, a set of components that activate for specific inputs and trigger the model behavior constitutes an internal mechanism.
Mechanistic interpretability techniques, such as activation maps applied to internal activations of the model~\cite{activation_visual} 
and logit lens~\cite{logit_lens}, allow the observation of model activations without modifying any parameters. Sparse autoencoders~\cite{sae} 
and task vectors~\cite{task_vectors} isolate distinct model components correlated to particular input concepts. Activation patching~\cite{activation_patching} tests causal hypotheses by replacing specific components to measure their direct impact on the signal flow. Causal testing manipulates internal signals such as activations, hidden states, and attention patterns of neural networks and evaluates through targeted interventions the resulting changes in model behavior~\cite{activation_patching}.  
These techniques have been extended to identify circuits~\cite{conmy2023towards, syed-etal-2024-attribution}, namely subgraphs within the computational graph of the neural network that are causally responsible for specific behaviors or functional properties of the model. 
% intended as a directed acyclic graph where nodes correspond to the operations performed during the forward pass and edges represent the activations flowing sequentially from one node to the next.
The computational graph is a directed acyclic graph where nodes correspond to the operations performed during the model forward pass, and edges represent the activations flowing sequentially from one node to the next.
Circuit discovery combines different mechanistic interpretability methodologies to find causal relationships between model components and shows the most promising results in terms of model interpretability~\cite{conmy2023towards}. 
While previous surveys on mechanistic interpretability~\cite{mi_survey_4, mi_survey_2026} mainly focus on mechanistic understanding within specific neural networks or emphasize the context of safety and robustness, we introduce a distinct perspective. We focus on how these diverse techniques are combined into interpretability frameworks to achieve a causal understanding of deep neural networks. Moreover, we provide examples of mechanistic methods for audio classification and image reconstruction.

\setlength{\tabcolsep}{6pt} % Adjust column spacing
\renewcommand{\arraystretch}{1.0} % Adjust row height
\begin{table}[t]
\centering
\caption{Comparison of traditional and mechanistic Interpretability.}
\resizebox{1.0\linewidth}{!}{%
\begin{threeparttable}
\begin{tabular}{m{4cm} m{7cm} m{7cm}}
\hline
 & \textbf{TRADITIONAL INTERPRETABILITY} & \textbf{MECHANISTIC INTERPRETABILITY} \\ 
\hline

\textbf{GOAL} & 
Identify input features that most contribute to the model final output & 
Identify model components that are causally responsible for the model behavior \\ 
\hline

\textbf{GRANULARITY} & 
Input-output relationship & 
Model inner mechanisms \\ 
\hline

\textbf{COMPUTATION COST} & 
Medium-low (generally a single forward pass and backward pass) & 
Medium-high (sparse autoencoder training, circuit discovery across all model components) \\ 
\hline

\textbf{SCALABILITY} & 
Medium-high (analysis on model input and final output) & 
Medium-low (causal testing iterating across all model components) \\ 
\hline

\textbf{LIMITATIONS} & 
Approximate, limited analysis of how the model produces internal activations & 
Poor performance under data distribution shift \\ 
\hline

\end{tabular}
\end{threeparttable}
}
\label{tab:xai_vs_mi}
\end{table}

\begin{figure}[t]
    \centering
    \includegraphics[width=\textwidth]{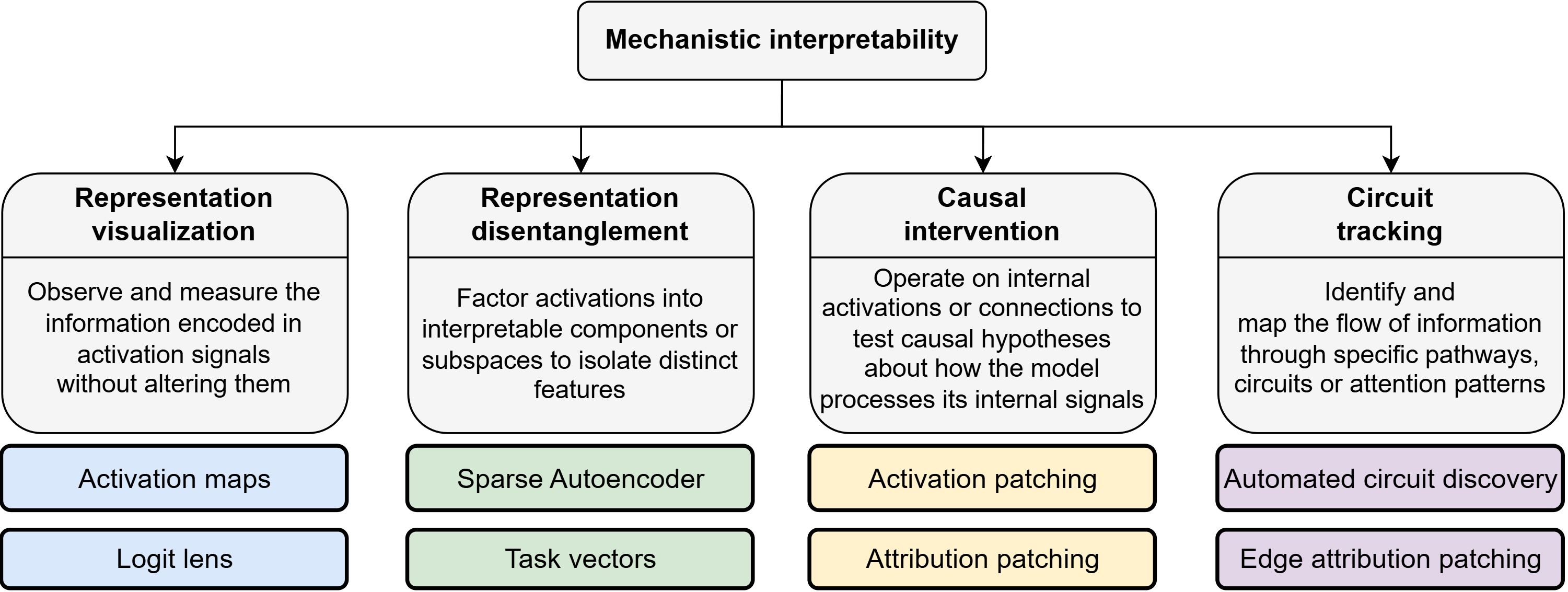}
    \caption{Categorization of mechanistic interpretability, along with representative methods under each category.}
    \label{fig:categories}
\end{figure}

We categorize techniques that support mechanistic interpretability frameworks (Figure~\ref{fig:categories}) into four groups: representation visualization, representation disentanglement, causal intervention, and circuit tracking. 
{\em Representation visualization} techniques observe and measure the information encoded in activation signals without altering them, through probes or visualization tools~\cite{activation_visual}. Techniques in this category generate hypotheses about which signals may contribute to specific model behaviors. {\em Representation disentanglement} factors activations into interpretable components or subspaces to isolate distinct features through dedicated methods such as sparse autoencoders~\cite{sae}.
Techniques in this category enable the identification of candidate neural variables that may serve as causal factors in the model internal computations.
{\em Causal intervention} operates on internal activations to test causal hypotheses about how the model processes its internal signals through intervention techniques such as activation patching~\cite{activation_patching}. Techniques in this category causally verify the relevance of specific components to the model computational flow. Finally, {\em circuit tracking} techniques identify and map the flow of information, finding specific pathways, circuits, or attention patterns in the network through, for example, automated circuit discovery~\cite{conmy2023towards} or edge attribution patching~\cite{syed-etal-2024-attribution}. Techniques in this category reconstruct causal chains linking internal activations to observed model behaviors.

% %%%%%%%%%%%%%%%%%%%%%%%%%%%%%%%%%%%%%%%%%%%%%%%%
\section*{Representation visualization}
\label{subsec:representation_visualization}

Representation visualization aims to observe and measure the activations without altering them~\cite{activation_visual}.
Activation maps are used to visualize the activations of model components. Unlike traditional interpretability techniques, which compute activation maps by considering the final output of the model, mechanistic approaches exploit these maps to measure responses of internal model components to specific inputs.
Visualizing the internal computations
is a preliminary step to understanding the model behavior~\cite{activation_visualization}. An activation map $A(\mathbf{x})$ is computed as the gradient of the activation $f(\mathbf{x}) \in \mathbb{R}^n$ with respect to the input $\mathbf{x} \in \mathbb{R}^d$, with $n>d$:
\begin{equation}
    A(\mathbf{x}) = \frac{\partial f(\mathbf{x})}{\partial \mathbf{x}}.
\end{equation}

\begin{figure}[t]
    \centering
    \includegraphics[width=0.75\textwidth]{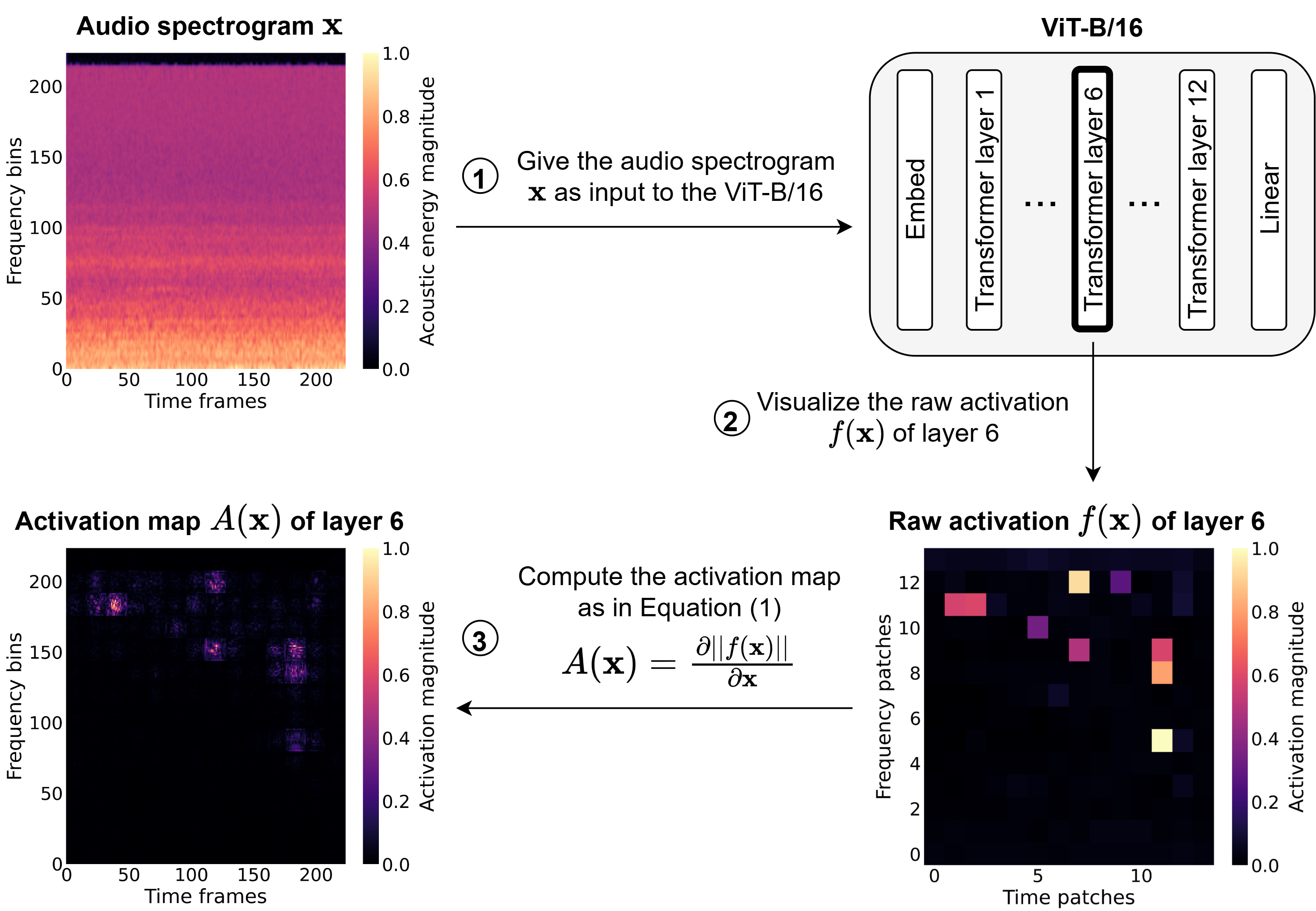}
    \caption{Activation map computation for ViT-B/16 audio spectrogram classification task at layer granularity. The activation map is computed for layer $6$.}
    \label{fig:example_activation_maps}
\end{figure}

Figure~\ref{fig:example_activation_maps} illustrates an end-to-end application of representation visualization using activation maps. We analyze a Vision Transformer (ViT-B/16), pretrained on ImageNet-1K and fine-tuned for $12$ epochs on the ESC-50 dataset~\cite{esc-50} for audio classification, achieving $88\%$ test accuracy. The one-dimensional audio signals from the ESC-50 dataset are transformed into spectrograms to serve as inputs to the model. In the input spectrogram $\mathbf{x}$, pixel intensity corresponds to acoustic energy magnitude. We extract the raw activation tensor $f(\mathbf{x})$ from layer $6$. The activation map $A(\mathbf{x})$ is obtained by computing the gradient of the raw activation $f(\mathbf{x})$ with respect to the input spectrogram $\mathbf{x}$. After normalizing the values of $A(\mathbf{x})$ to $[0, 1]$, the resulting map highlights the regions of the input that most influence the internal state of the layer. The highlighted regions are the time-frequency bins where small perturbations would cause the most substantial changes in the output of layer $6$. $A(\mathbf{x})$ reveals that layer $6$ localizes specific acoustic events occurring in mid-to-high frequency bands (bins $130-200$) at distinct time frames.

In addition to activation maps, which visualize the model prediction layer by layer, the logit lens~\cite{logit_lens} converts the logits of hidden states into the output space of the model. This is useful to observe how the model processes its input till the final prediction.
Representation visualization methods have been exploited to provide descriptions of neuron activations~\cite{activation_visualization}. By collecting the images that most activate the neuron of interest, captions for such images are generated, along with images
corresponding to each generated caption. The neuron is described by selecting the best caption based on the corresponding image on which the neuron activates the most.
Observing model activations reveals that scaled models are not easier to interpret than original models~\cite{zimmermann2023scale}. The process involves the selection of random neurons in the scaled and non-scaled models and visualizing the corresponding activations. Then, some participants are presented with a pair of strongly and weakly activating images for the target neuron and are asked to identify the strongly activated one. Participants showed higher identification accuracy when evaluating the non-scaled neural network.
In the context of neural network alignment, investigating the model internal mechanisms ensures harmlessness following safety fine-tuning by visualizing neuron activations within transformer blocks~\cite{jain2024makes}. This specific training stage aims to mitigate toxic behavior without degrading overall performance capabilities. When presented with a malicious input, an aligned model must recognize the hazard and modulate its output accordingly. By comparing internal representations triggered by benign versus harmful prompts, the fine-tuning process separates safe and unsafe activations.

Representation visualization methods identify the internal mechanisms of Low-Rank Adaptation (LoRA) fine-tuning within neural networks for speech emotion recognition~\cite{sp_application_2}. This analysis leverages the logit lens, which inspects the network by projecting intermediate layer representations into the final classification space. The representation extracted from a given layer is passed through the model to generate intermediate logits. By evaluating these intermediate predictions against the final output using the Kullback-Leibler (KL) divergence, it is possible to show that early encoder layers preserve general
features, while deeper layers contain the most significant contributions of LoRA.

% %%%%%%%%%%%%%%%%%%%%%%%%%%%
\section*{Representation disentanglement}
\label{subsec:representation_disentanglement}

A neural network can learn more concepts than its number of neurons by encoding features in overlapping activation subspaces. This phenomenon is known as the superposition hypothesis~\cite{superposition}: while this dense representation increases model capacity, it introduces polysemanticity. Polysemanticity is the phenomenon where a single neuron responds to multiple concepts~\cite{polysemanticity}. A polysemantic neuron 
confounds the causal pathways of all the distinct concepts it simultaneously encodes, thus making it difficult to perform interventional tests. 
This behavior is a major obstacle to causality, since the same activations can reflect several patterns depending on context, thus preventing a direct mapping between neurons and input concepts. A polysemantic network can be conceptualized as a highly compressed approximation of a much larger monosemantic model, where each neuron acts as a dedicated channel activating exclusively for a single input feature.

\begin{figure}[t]
    \centering
    \includegraphics[width=0.65\textwidth]{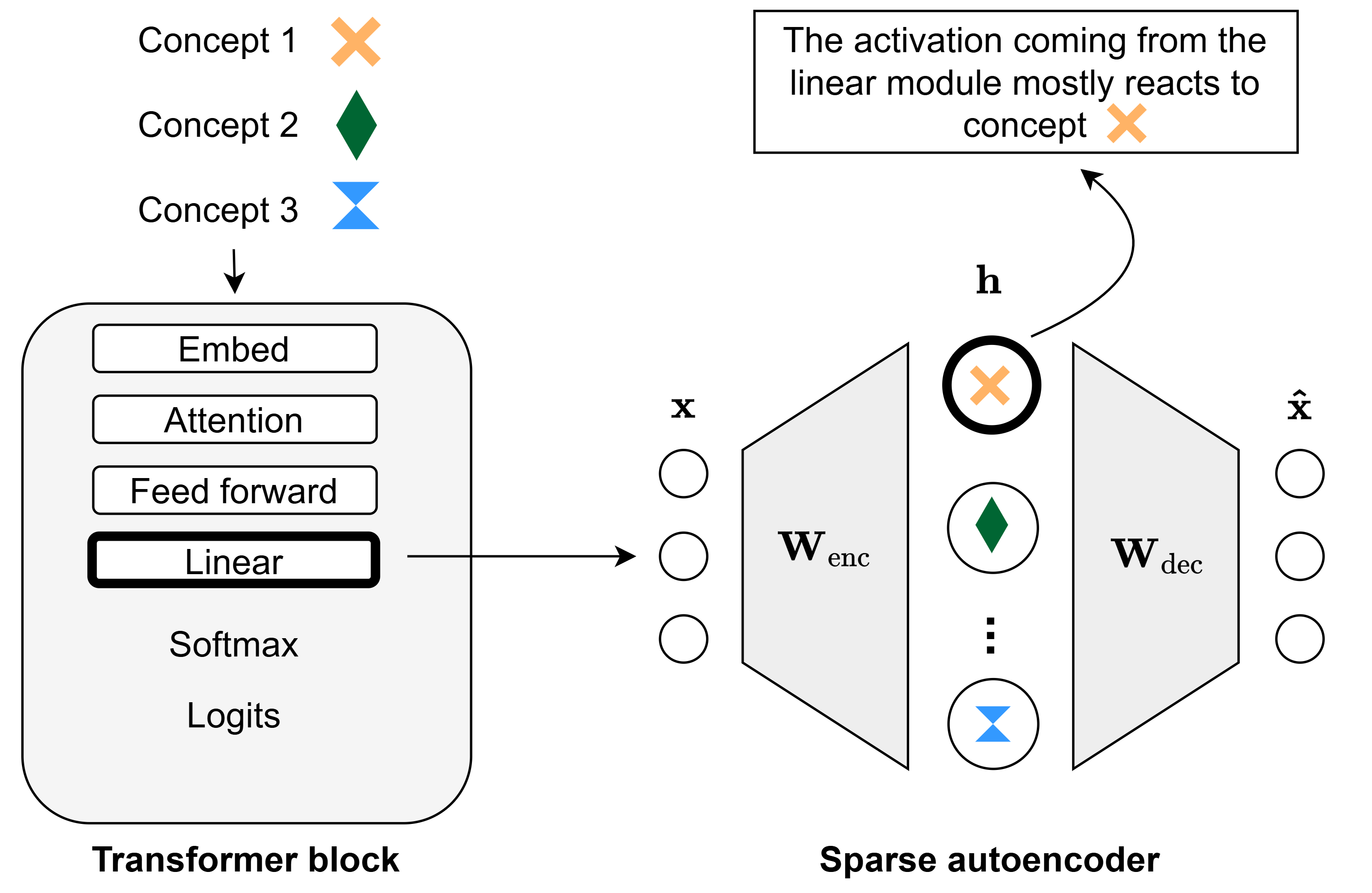}
    \caption{Sparse autoencoder applied to the linear module activation of a transformer block at module granularity. Each monosemantic feature in the hidden space of the sparse autoencoder reacts to the input concept it is associated with.}
    \label{fig:sae}
\end{figure}

Disentanglement is the process of converting a polysemantic signal into a monosemantic component and producing monosemantic representation spaces. Monosemanticity is the phenomenon where each neuron corresponds to a single concept. To interpret deep neural networks and deal with the superposition hypothesis~\cite{superposition}, internal neural components must be disentangled into interpretable monosemantic components or subspaces to isolate distinct features.
Upon this, task vectors~\cite{task_vectors} are linear directions within a neural network parameter space that isolate discrete learned behaviors. Defined as the difference between the weights of a task-specific fine-tuned model and its pre-trained base model, these vectors embed input concepts as nearly orthogonal separable features within a high-dimensional continuous space.
Sparse autoencoders can be used to project polysemantic activations into a monosemantic space~\cite{sae}. A sparse autoencoder takes as input an activation $\mathbf{x} \in \mathbb{R}^d$ of a model component. The encoder $\mathbf{W}_{\text{enc}} \in \mathbb{R}^{n \times d}$ projects $\mathbf{x}$ into a high-dimensional space $\mathbb{R}^n$, where $n>d$, and computes the monosemantic features activations $\mathbf{h} \in \mathbb{R}^n$:
\begin{equation}
    \mathbf{h} = \text{ReLU}(\mathbf{W}_{\text{enc}} \mathbf{x} + \mathbf{b}_{\text{enc}})
\label{eq:sae_1}
\end{equation}
where $\mathbf{b}_{\text{enc}}$ is the encoder bias, and $\text{ReLU}(\mathbf{x})=\text{max}(0,\mathbf{x})$ is the Rectified Linear Unit function ensuring that any negative values become zero. The decoder $\mathbf{W}_{\text{dec}} \in \mathbb{R}^{d \times n}$ reconstructs the original activation $\mathbf{\hat{x}} \in \mathbb{R}^d$ from the monosemantic features:
\begin{equation}
    \mathbf{\hat{x}} = \mathbf{W}_{\text{dec}}\mathbf{h} + \mathbf{b}_{\text{dec}}
\label{eq:sae_2}
\end{equation}
where $\mathbf{b}_{\text{dec}} \in \mathbb{R}^d$ is the decoder bias. The loss function $\mathcal{L}$ to be minimized is:
\begin{equation}
    \mathcal{L} = ||\mathbf{x}-\mathbf{\hat{x}}||^2_2 + \lambda||\mathbf{h}||_1
\label{eq:sae_3}
\end{equation}
where $||\mathbf{x}-\mathbf{\hat{x}}||^2_2$ is the mean squared error that forces the sparse autoencoder to accurately reconstruct the original activation, and $\lambda||\mathbf{h}||_1$ is the regularization term forcing the reconstruction to use as few monosemantic features as possible. The reconstructed input $\mathbf{\hat{x}}$ is exploited to train the sparse autoencoder and compute the monosemantic features $\mathbf{h}$.
% (see Figure~\ref{fig:sae}).
Figure~\ref{fig:sae} shows the functionality of a sparse autoencoder, whose monosemantic features activate based on the input concepts.

Sparse autoencoders have been employed to ensure that representations learned by foundation models align with the ontological structure of the ImageNet dataset~\cite{Olson_2025_CVPR}. This ontology is organized in sets of synonymous terms that represent a single concept. The sparse autoencoder is trained on the activation of the foundation model layers. The activation maps computed on the hidden states of the sparse autoencoder with respect to the input images highlight input features corresponding to hierarchical concepts, confirming that the model effectively reveals the hierarchical structure.
Furthermore, sparse autoencoders have been adopted in diffusion models for concept unlearning~\cite{cywinski2025saeuron}. By disentangling model components and analyzing the corresponding activations under specific concepts, activations responsible for undesired concepts are scaled with a negative multiplier at inference time to control the generation process.
Similarly, uncovering internal mechanisms of bias generation in diffusion models supports bias mitigation~\cite{Shi_2025_CVPR}. Representation disentanglement provides the identification of neurons correlated with bias generation. In this way, it is possible to control the bias level in the generated content by adjusting the scale of such identified neurons.
Moreover, to disentangle convolutional neural networks, inserting middle layers with monosemantic channels within the model allows for the decomposition of polysemantic channels into monosemantic ones. By tuning the parameters of the added disentangled channels at training time, the new network still has comparable performance to the original one~\cite{Hesse_2025_CVPR}.

Training a sparse autoencoder requires the activations extracted from the specific model components under analysis, depending on the chosen granularity. 
Given that the encoder projects the signal into a higher dimensional space, 
the sparse autoencoder training is generally limited to the considered model activations~\cite{Shi_2025_CVPR}. 
However, there is no guarantee that the high-dimensional space of a sparse autoencoder will contain the specific concept to be analyzed. 
In applications such as concept unlearning~\cite{cywinski2025saeuron} or bias mitigation~\cite{Shi_2025_CVPR}, whose aim is to penalize activations responsible for specific concepts or bias in the neural networks, it becomes particularly challenging to penalize components related to undesired model behaviors if those concepts are absent from the learned representations of the sparse autoencoder.

\section*{Causal intervention}
\label{subsec:causal_intervention}

Causal intervention techniques modify activations to verify their contribution to the internal processing mechanisms.
To determine the causal impact of a specific component on the model behavior, activation patching~\cite{activation_patching} is a causal tool that relies on ablation to measure whether the model output changes accordingly. If the prediction shifts, the patched activation is deemed causally responsible for that behavior.

Given the activation $f(\mathbf{x}) \in \mathbb{R}^n$ of a model component $f$ when the input is $\mathbf{x} \in \mathbb{R}^d$, with $n > d$, the most used ablation techniques are to replace $f(\mathbf{x})$ with a zero-tensor of the same dimension, $f(\mathbf{x}) \leftarrow 0$ (zero ablation); with its average value across a dataset $\mathcal{D}$, $f(\mathbf{x}) \leftarrow \mathbb{E}_{x \sim \mathcal{D}}[f(\mathbf{x})]$ (mean ablation); or with its version coming from a different input $\mathbf{\tilde{x}}$ to test which part of the network carries specific information, 
$f(\mathbf{x}) \leftarrow f(\mathbf{\tilde{x}}),$ (interchange ablation). 

\begin{figure}[htbp]
    \centering
    \subfloat[\label{fig:patching_diagram}]{%
        \includegraphics[width=0.8\linewidth]{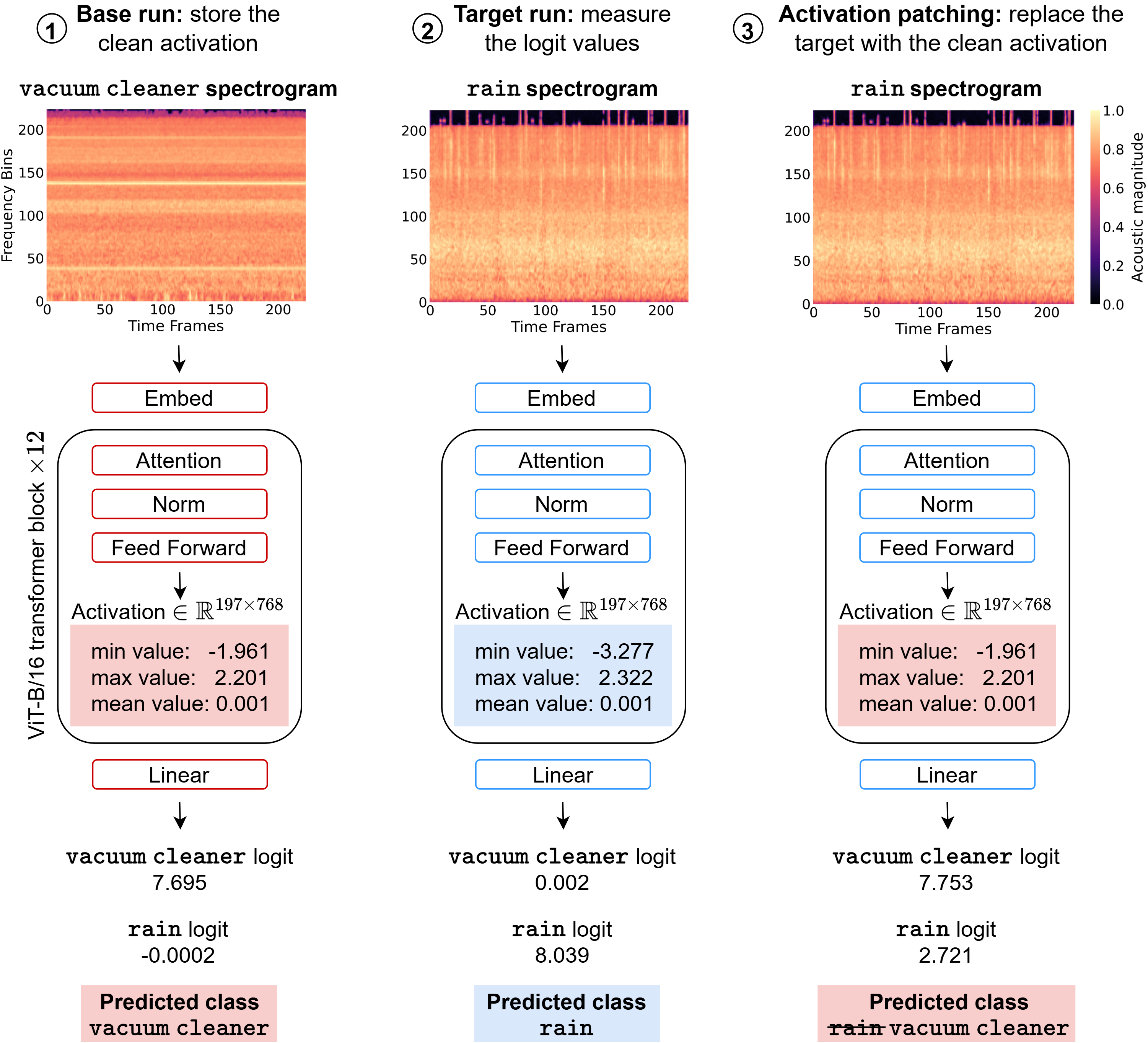}%
    }

    \subfloat[\label{fig:patching_curve}]{%
        \includegraphics[width=0.37\linewidth]{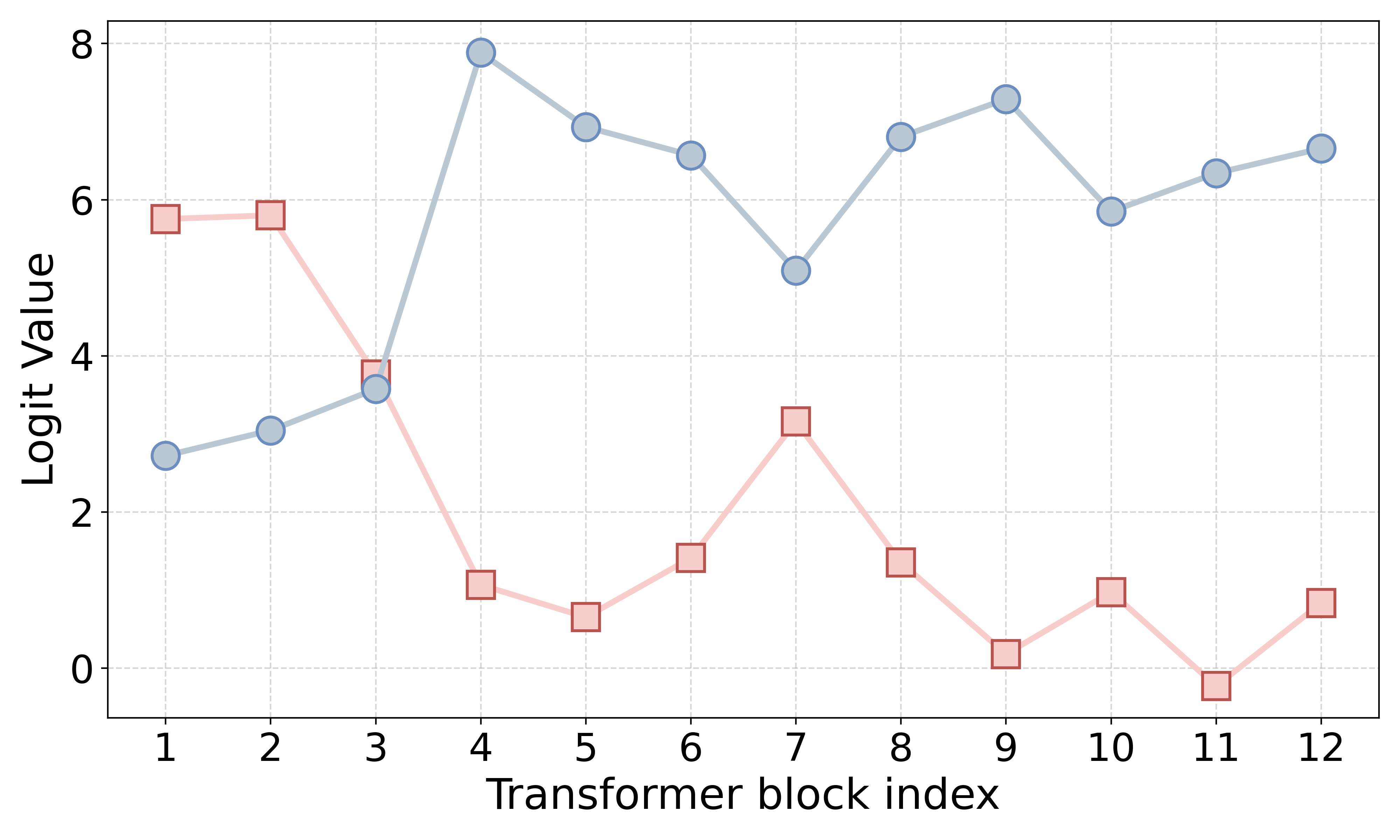}%
    }
    \caption{Activation patching for ViT-B/16 audio spectrogram classification at module granularity. (a) Interchange ablation on the feed forward module of the first transformer block. Injecting the base run activation into the target run alters the logits and restores the \texttt{vacuum cleaner} prediction. (b) Logits for \texttt{vacuum cleaner} (\basesquare) and \texttt{rain} (\targetcircle) during activation patching applied to every feed forward module separately. Patching any of the first three feed forward modules raises the \texttt{vacuum cleaner} logit, reverting the prediction from \texttt{rain}.}
    \label{fig:example_activation_patching}
\end{figure}

Figure~\ref{fig:example_activation_patching} reports an end-to-end application of activation patching via interchange ablation for audio spectrogram classification, utilizing the same ViT-B/16 model with the ESC-50 dataset~\cite{esc-50} used for Figure~\ref{fig:example_activation_maps}. In the base run, the model processes and correctly classifies a \texttt{vacuum cleaner} spectrogram. In the target run, it correctly classifies a \texttt{rain} spectrogram.
We apply activation patching to identify which feed forward modules within the transformer blocks are causally responsible for the \texttt{vacuum cleaner} prediction. To this end, we perform a causal intervention during the target run, replacing the feed forward activation with the corresponding activation saved from the base run. We then observe the resulting changes in the output logits and the final predicted class. If this intervention causes the \texttt{vacuum cleaner} logit to increase enough to change the model prediction, we can conclude that the patched activation drives the model toward the base class.
As depicted in the third step of Figure~\ref{fig:example_activation_patching}(a), patching the feed forward module in the first transformer block causes the \texttt{vacuum cleaner} logit in the target run to increase from $0.002$ to $7.753$, restoring the base prediction. Furthermore, Figure~\ref{fig:example_activation_patching}(b) details the logit values for both classes when this intervention is applied independently across the feed forward module of each block. The \texttt{vacuum cleaner} logits surpass the \texttt{rain} logits when patching any of the first three feed forward modules, showing that these early layers are causally responsible for the \texttt{vacuum cleaner} prediction.

Activation patching with interchange ablation requires a separate forward pass for every component intervened upon. Attribution patching~\cite{attribution_patching} provides a computationally efficient approximation of the causal effect of an activation on a performance measure $M$. The complete procedure to apply attribution patching is detailed in Algorithm~\ref{alg:attribution_patching}. Attribution patching requires two forward passes (one to store the activations from the base run and one from the target run) and a single backward pass to compute the model gradient.
Attribution patching estimates the change in the measure, denoted as the scalar $\Delta M$, using a first-order Taylor expansion around the base activation. Let $f(\mathbf{x}) \in \mathbb{R}^n$ denote the activation of a model component $f$ given a base input $\mathbf{x}$, and $f(\tilde{\mathbf{x}}) \in \mathbb{R}^n$ denote the activation given a target input $\tilde{\mathbf{x}}$. Let $M: \mathbb{R}^n \to \mathbb{R}$ be a scalar measure evaluating the model output, such as loss or logit difference. Attribution patching estimates the effect of intervening on $f$ as:
\begin{equation}
    \Delta M = \frac{\partial M}{\partial f(\mathbf{x})} \cdot (f(\tilde{\mathbf{x}}) - f(\mathbf{x})),
\label{eq:attribution_patching}
\end{equation}
\noindent where $\frac{\partial M}{\partial f(\mathbf{x})} \in \mathbb{R}^n$ is the gradient of the performance measure with respect to the base activation, and the $\cdot$ operator denotes the dot product over all dimensions of the activation.

Causal intervention verifies whether internal model components function as causal mediators between input data and model outputs. In vision-language models, causal interventions have been used to determine the cause-effect relationship between specific visual tokens and object identification accuracy~\cite{neo2025towards}. After representation visualization identifies candidate visual tokens correlated with a target concept, counterfactual intervention via mean ablation is performed on those tokens. If model performance degrades when the activation of these components is altered, their causal responsibility in producing the final prediction is established.
In diffusion models, cause-effect hypotheses on feature disentanglement are validated using interchange ablation~\cite{Takatsuki_2025_CVPR}. By replacing internal representations with corrupted activations derived from perturbed inputs, researchers evaluate whether restoring specific disentangled features causally steers the generated output toward the target distribution.
Furthermore, causal intervention can identify the internal mechanisms responsible for behavioral safety triggers in vision-language models exposed to toxic inputs~\cite{xu2025crossmodal}. When facing a toxic input, if the safety mechanism is activated, the model refuses the instructions and apologizes to the user. Therefore, a reply with an apology is a signal of the safety mechanism. Based on that, selecting samples that successfully enable the safety mechanism enables the analysis of the triggered model
activations. Layers responsible for the safety mechanism exhibit attention peaks when the input is toxic. This is validated by masking the toxic information and testing the changes in the safety capabilities of the neural network.

\begin{algorithm}[t]
\small
\caption{Attribution Patching}
\label{alg:attribution_patching}
\begin{algorithmic}[1]

\renewcommand{\algorithmicrequire}{\textbf{Input:}}
\renewcommand{\algorithmicensure}{\textbf{Output:}}

\Require Base input $\mathbf{x}$, Target input $\tilde{\mathbf{x}} \in \mathbb{R}^d$, Model component $f$, Scalar performance measure $M$
\Ensure Estimated change in measure $\Delta M$ 

\State \textbf{Forward Passes:} Store base activation $f(\mathbf{x}) \in \mathbb{R}^n$, base measure $M$, and target activation $f(\tilde{\mathbf{x}}) \in \mathbb{R}^n$

\State \textbf{Backward Pass:} $\mathbf{g} \leftarrow \frac{\partial M}{\partial f(\mathbf{x})}$ \Comment{ Compute gradient w.r.t base activation}

\State \textbf{Attribution:} $\Delta M \leftarrow \mathbf{g} \cdot (f(\tilde{\mathbf{x}}) - f(\mathbf{x}))$ \Comment{Compute linear approximation}

\State \Return $\Delta M$
\end{algorithmic}
\end{algorithm}

% %%%%%%%%%%%%%%%%%%%%%%%%%%%
\section*{Circuit tracking}
\label{subsec:circuit_tracking}

Circuit tracking aims to identify model components and connections, and reconstruct circuits that link internal activations to observed behaviors~\cite{conmy2023towards}.

Model components are sequentially subjected to causal interventions, such as zero, mean, or interchange ablation, to verify their contribution to a specific model behavior and discover circuits. Unlike causal intervention techniques where components are checked independently, circuit tracking iteratively applies interventions to identify a circuit. Automated circuit discovery~\cite{conmy2023towards}, whose procedure is detailed in Algorithm~\ref{alg:acdc}, simplifies neural networks by applying activation patching to the edges of the model computational graph, which correspond to the activations passed between network components. Processing the graph in reverse topological order, the algorithm removes edges whose ablation only marginally degrades the model output. The removal criterion is based on the KL divergence between the original model prediction and the prediction from the ablated circuit. Let $C_{i-1}$ be the circuit at the previous iteration $i-1$, and $C_{i-1}\setminus \{f_i\}(\mathbf{x})$ denote the output distribution of the circuit when the activation $f_i(\mathbf{x}) \in \mathbb{R}^n$ of component $f_i$ is ablated. For an input $\mathbf{x} \in \mathbb{R}^d$ and the predefined threshold $\tau > 0$, the circuit is updated as follows:
\begin{equation}
    C_i = \begin{cases} 
    C_{i-1} \setminus \{f_i\} & \text{if } D_{KL}(F(\mathbf{x}) \parallel C_{i-1}\setminus \{f_i\}(\mathbf{x})) < \tau \\ 
    C_{i-1} & \text{otherwise} 
    \end{cases}
    \quad \text{for } i=1, \cdots, E,
\label{eq:acdc}
\end{equation}
\noindent where $F(\mathbf{x})$ is the output distribution of the original neural network and $E$ is the number of edges.

\begin{algorithm}[t]
\small 
\caption{Automated Circuit Discovery}
\label{alg:acdc}
\begin{algorithmic}[1] 
\renewcommand{\algorithmicrequire}{\textbf{Input:}}
\renewcommand{\algorithmicensure}{\textbf{Output:}}

\Require Model $F$, input $\mathbf{x} \in \mathbb{R}^d$, threshold $\tau > 0$, components $\{f_1, \dots, f_E\}$ in reverse topological order
\Ensure Circuit $C$

\State $C_0 \leftarrow F$ \Comment{Initialize the circuit}
\For{$i=1$ to E}
    \State $C_{i} \leftarrow C_{i-1} \setminus \{f_i\}$ \Comment{Temporarily remove $f_i$}
    \State $D \leftarrow D_{KL}(F(\mathbf{x})||C_{i}(\mathbf{x}))$ \Comment{Compute the KL Divergence}
    \If{$D<\tau$}
        \State Keep $f_i$ removed from $C_i$
    \Else
        \State $C_i \leftarrow C_{i-1}$ \Comment{Retain $f_i$}
    \EndIf
\EndFor
\State \Return $C \leftarrow C_E$ 
\end{algorithmic}
\end{algorithm}

However, iterating over all the components of a deep neural network requires high computational time, especially when executing interchange ablation, where a base and a target run are required for each iteration (Figure~\ref{fig:example_activation_patching}). Therefore, edge attribution patching~\cite{syed-etal-2024-attribution} aims to discover circuits by employing attribution patching~\cite{attribution_patching} across all the model activations and sorting them in descending order based on the $\Delta M$ scores (Equation~\eqref{eq:attribution_patching}). The top $k$ activations will be part of the final circuit. Figure~\ref{fig:example_eap} illustrates an end-to-end application of edge attribution patching with zero ablation for the image reconstruction task. We analyze a convolutional autoencoder with $6$ layers, which was trained for $15$ epochs on the CIFAR-10 dataset, achieving a peak signal-to-noise ratio of $24$ and Mean Squared Error (MSE) of $0.004$. The images from CIFAR-10 are perturbed by exploiting zero-mean Gaussian noise with standard deviation of $0.15$. As depicted in Figure~\ref{fig:example_eap}(a), we report the minimum, maximum, and mean values of the first layer activation, although $\Delta M$ is estimated for every layer. We save all the activations and the performance measure $M$ during the forward pass. We use the MSE as the measure $M$. We adopt zero ablation and replace the corrupted activation with a zero tensor of the same dimension, and compute the $\Delta M$ score as in Equation~\eqref{eq:attribution_patching}. In Figure~\ref{fig:example_eap}(b), we plot the estimated $\Delta M$ for each layer. For this experiment, we choose to keep the top-3 layers in the circuit responsible for the reconstructed image, since the scores for these layers are much higher than those of the other layers. Therefore, the final circuit is made up of layer $1$ ($\Delta M = 0.00058$), layer $5$ ($\Delta M = 0.00059$), and layer $6$ ($\Delta M = 0.00089$).

\newsavebox{\tallimagebox}

\begin{figure}[t]
    \centering
    
    \sbox{\tallimagebox}{\includegraphics[width=0.59\linewidth]{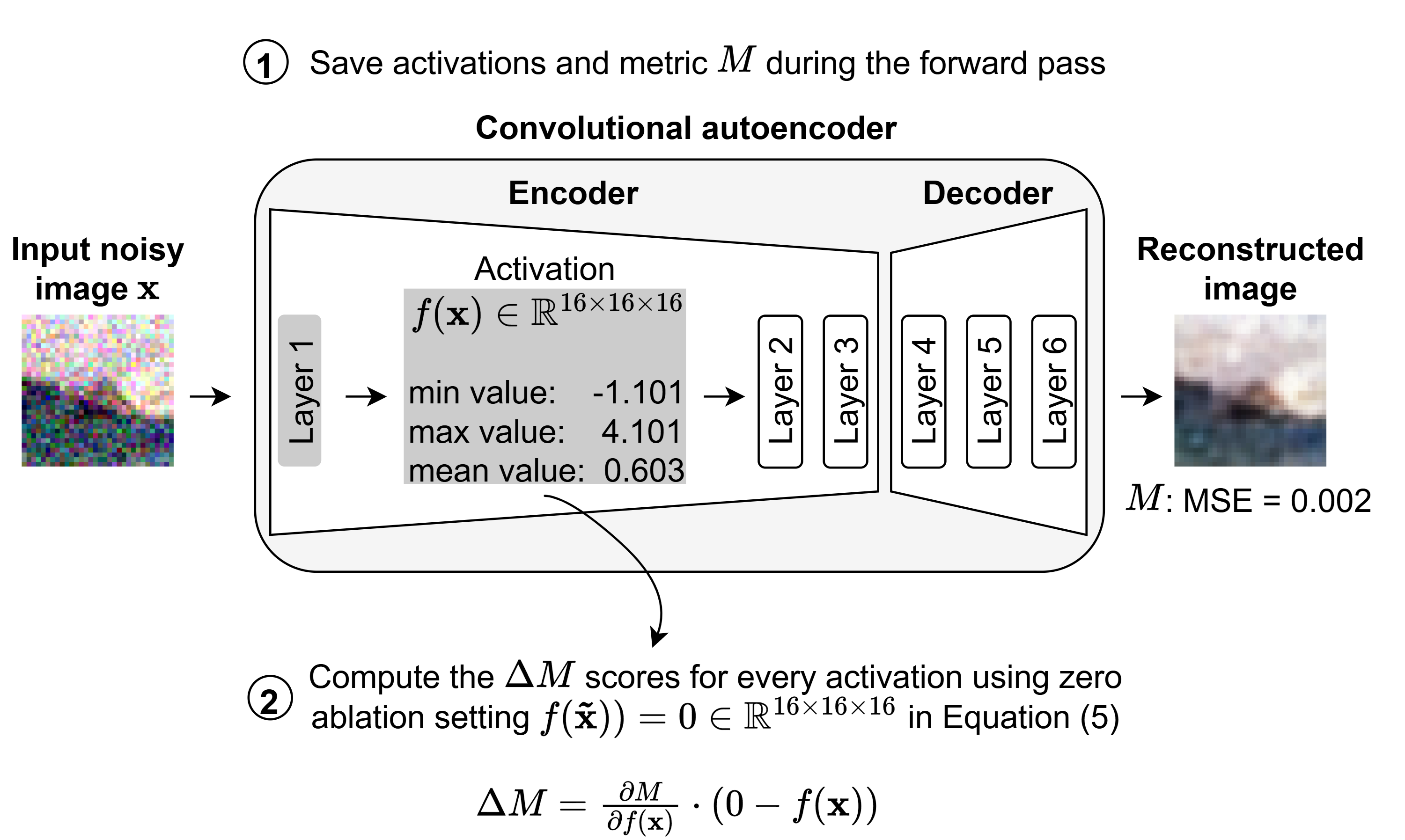}}
    
    \subfloat[\label{fig:eap_diagram}]{%
        \usebox{\tallimagebox}%
    }
    \hfil
    \subfloat[\label{fig:attribution_scores}]{%
        \parbox[b][\ht\tallimagebox][c]{0.39\linewidth}{%
            \centering
            \includegraphics[width=\linewidth]{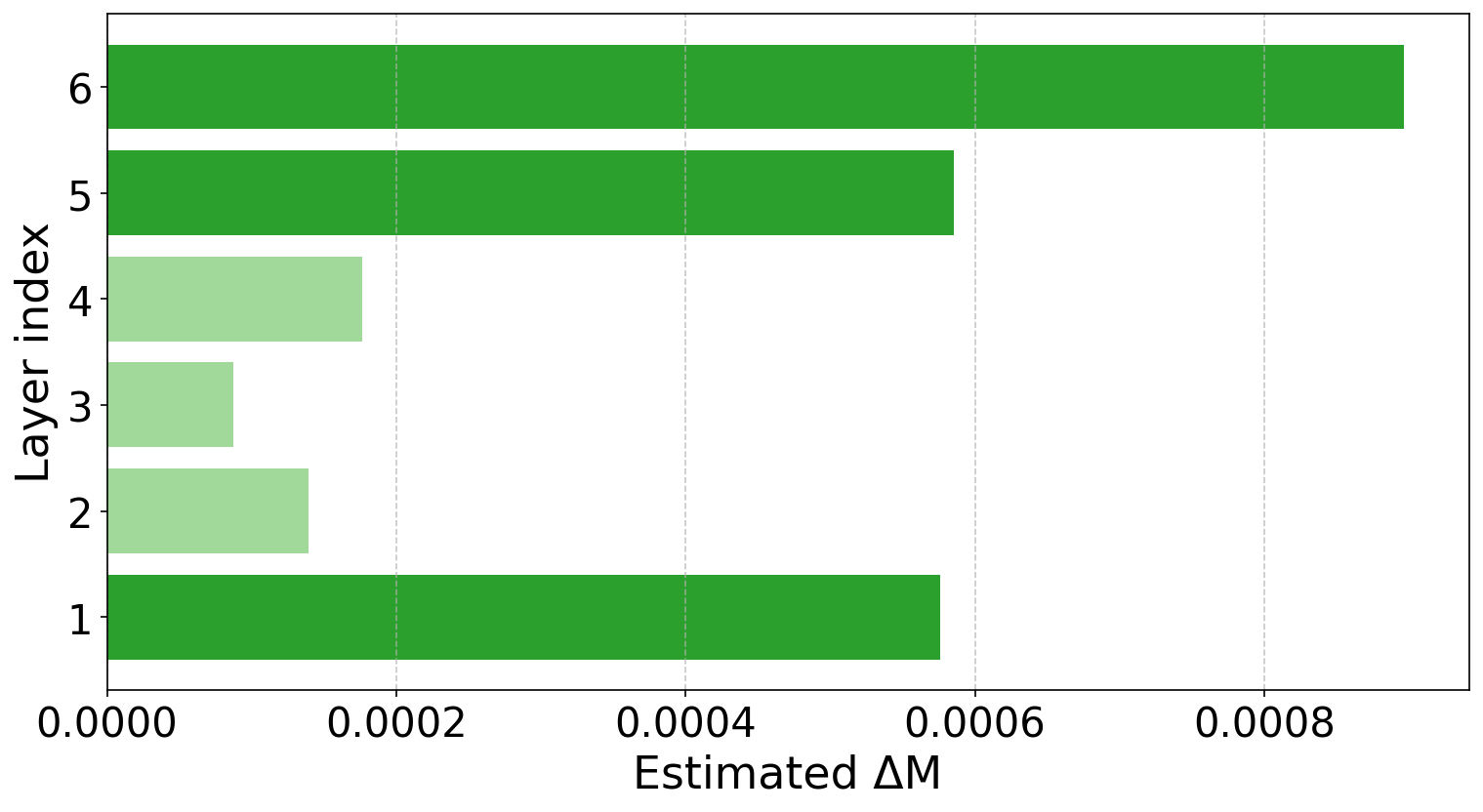}%
        }%
    }
    \caption{Edge attribution patching with zero ablation for the convolutional autoencoder at layer granularity. (a) Estimation of $\Delta M$ for the first encoder layer. The performance metric $M$ is the Mean Squared Error (MSE). A scalar $\Delta M$ score is obtained by summing all elements of the resulting $\Delta M$ tensor. (b) $\Delta M$ for each layer. The top-$k$ layers (\textcolor[HTML]{2ca02c}{\rule{1.2ex}{1.2ex}}) with the highest $\Delta M$ scores ($k=3$ is chosen for this implementation) constitute the circuit responsible for reconstructing the input image. The remaining layers (\textcolor[HTML]{a1d99b}{\rule{1.2ex}{1.2ex}}) are excluded from the circuit.}
    \label{fig:example_eap}
\end{figure}

Beyond empirical extraction, circuit tracking can establish lower bounds on model performance by ablating specific attention matrices to infer the causal importance of individual tokens~\cite{gross2024compact}. 
Furthermore, circuit tracking analyzes whether neural networks can generalize across data distribution shifts~\cite{wang2024grokking}. Activation patching is applied to verify if the target state is altered when providing the model with the perturbed input. If the state is altered, then it will be part of the final circuit made up of nodes and connections responsible for generalization. Interestingly, models trained with grokking, consisting of an extended training far beyond overfitting~\cite{wang2024grokking}, have higher generalization performance to data unseen during training than models trained for fewer epochs.

\section*{Interpretability frameworks}
\label{subsec:mixed_methods}

\begin{figure}[t]
    \centering
    \includegraphics[width=0.7\textwidth]{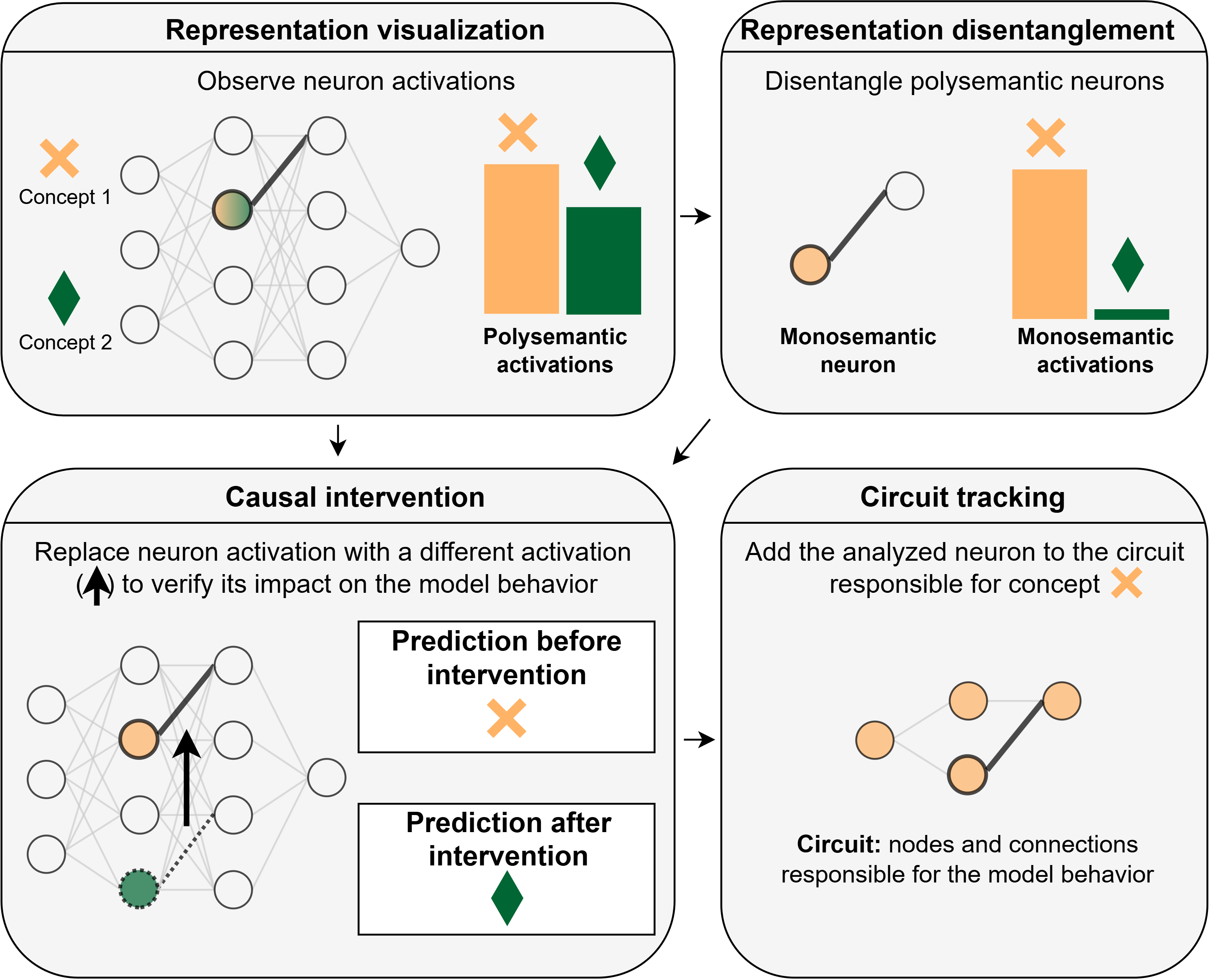}
    \caption{
    Combining mechanistic interpretability categories at neuron granularity: activations are observed through visualization, isolated through disentanglement, tested for behavioral impact through causal intervention, and are part of the circuit responsible for the model behavior if the causal intervention on that activation changes the prediction.
    }
    \label{fig:approaches_interaction}
\end{figure}

Frameworks combine interpretability methods to observe model activations, project them into monosemantic spaces, test the causal importance of these isolated features through targeted amplification or suppression, and track the interconnected pathways to identify circuits responsible for a particular behavior. This can be used to steer the model towards target behaviors, such as removing concepts that contribute to harmful consequences, reinforcing stereotypes, and exacerbating inequalities in various social contexts~\cite{Shi_2025_CVPR}. As shown in Figure~\ref{fig:approaches_interaction}, mechanistic interpretability categories interact within frameworks. Model activations observed through representation visualization are disentangled and tested through causal intervention. Then, if the causal intervention reveals such an activation as responsible for the model behavior, it will be part of the related circuit.

Certain interpretability frameworks  {\em bypass explicit feature disentanglement}, directly linking representation visualization to causal intervention~\cite{Takatsuki_2025_CVPR, neo2025towards, xu2025crossmodal}. Rather than relying on monosemantic projection, these methods directly verify the functional relevance of specific components, such as vision transformer attention heads, by replacing them with corrupted activations derived from noisy inputs~\cite{Chen_2025_NeurIPS, Li_2026_AAAI}. While replacing an activation with the version from another forward pass enables the impact measurement of a component on model behavior, steering the corrupted activation toward its clean state while patching subsequent activations with the corresponding corrupted activations helps to visualize the component activation~\cite{Takatsuki_2025_CVPR}. By passing such an isolated activation to an image generator, it is possible to produce an image containing the concepts learned by that component.

Linear probing and activation patching can be used to uncover the internal mechanisms of automatic speech recognition models~\cite{sp_application_1}. By training linear probes on intermediate activations, encoders for speech recognition build semantic representations, distinguishing between word categories with $85.6\%$ accuracy by the final encoder layer. Probing the decoder residual stream enables the prediction of hallucinations with up to $93.4\%$ accuracy. To investigate acoustic-contextual trade-offs, activation patching is used to inject white noise into activations of the encoder. This intervention restores acoustic accuracy in $85\%$ of error cases.

Pure representation visualization is fundamentally constrained by polysemanticity~\cite{polysemanticity}. Because individual neurons frequently encode multiple, superimposed concepts, activation maps often yield ambiguous interpretations. Consequently, while representation visualization is effective for generating causal hypotheses about internal signals, verifying these hypotheses requires transitioning to active interventional techniques and representation disentanglement. Other interpretability frameworks {\em rely on representation disentanglement} combined with {\em causal intervention} to dissect and control model behavior~\cite{Shi_2025_CVPR, cywinski2025saeuron, Chen_2026_AAAI}.
These configuration maps embed polysemantic activations into isolated monosemantic features to identify the ones that drive specific behaviors. Subsequently, scaling the underlying signal amplitudes facilitates the mitigation of undesirable outputs.
Because interventional signal modeling validates whether hypothesized monosemantic features are necessary components of the model internal processing mechanisms, the causal intervention approach is coupled with circuit tracking to construct reliable circuits~\cite{conmy2023towards, gross2024compact, gupta2024interpbench, syed-etal-2024-attribution}. 
Furthermore, to build these causal chains from the ground up, representation visualization frequently precedes both causal intervention and circuit tracking to preliminarily detect candidate signals of interest~\cite{wang2024grokking}. This comprehensive integration allows candidate features to be isolated via initial visual probing and tested through interventional ablation.

%%%%%%%%%%%%%%%%%%%%%%%%%%%%%%%%%%%%%%%%%%%%%%%%%%%%%%%%%%%%%%%%%%%%%%
\section*{Perspectives}
\label{sec:open_problems}

Based on the above discussions, we finally present open challenges in causality research for deep neural networks, which include generalization, alignment of internal representations with human-interpretable concepts, model robustness, efficiency, and sensitivity.

\subsection*{Generalization}

Mechanistic interpretability may help address the performance degradation of deep neural networks when employed in out-of-distribution scenarios~\cite{computer_vision_application}. This generalization problem occurs when deep models learn dataset artifacts to minimize the loss function. Through targeted causal interventions, model activations that respond when the model is employed in out-of-distribution scenarios can be identified, constraining the forward pass to rely exclusively on these causal subgraphs. Such identified activations can also be rewarded at training time through tailored loss regularizations, steering the model to rely on the internal components that increase generalization performance. This would prevent the network from exploiting training data artifacts, providing better performance on out-of-distribution scenarios.

\subsection*{Alignment of internal representations
with human-interpretable concepts}

While representation disentanglement techniques can produce monosemantic components~\cite{sae, Hesse_2025_CVPR}, these extracted signals may lack semantic grounding~\cite{ji}. To map model components to human-understandable concepts, structured concept supervision could be integrated to steer the discovery of interpretable features. To validate these mappings and ensure that an isolated signal aligns with a human concept, causal interventions could then be integrated with human-in-the-loop feedback or automated concept probers~\cite{activation_visual}.

\subsection*{Robustness}
Certified robustness techniques aim to provide the accuracy lower bound of a model against any adversarial attack in a predefined perturbation region of the input~\cite{randomized_smoothing}. Interval bound certification methods propagate this perturbation region layer by layer~\cite{ibp}. Since they require a single forward pass, these methods are computationally efficient but suffer from propagation errors on the estimated intervals, resulting in inaccurate bounds as network depth increases. 
Randomized smoothing methods can deliver accurate certifications for deeper architectures, but are characterized by high computational cost due to the extensive number of forward passes required to produce the certification for a single input~\cite{randomized_smoothing}.
Mechanistic interpretability potentially supports a certification process by isolating specific model components responsible for certification failure. Causal intervention techniques can map the propagation of adversarial noise through the neural network and isolate the specific model components that amplify malicious signals. Targeted interventions can be applied to these vulnerable components, improving the model certified robustness without incurring the high computational costs of smoothing the entire network.

\subsection*{Efficiency}

Model pruning identifies components of a deep neural network with low impact on accuracy~\cite{scalability_sp_letters}. To isolate prunable components, traditional magnitude-based methods remove trainable parameters relying on their norm. Although such methods can yield significant reductions in parameter count and computational overhead, performance may degrade accordingly. Causal intervention can help analyze such causal impact to remove only the activation of a component identified as prunable. This would facilitate the generation of architecturally efficient models. For example, automated circuit discovery~\cite{conmy2023towards} removes edges in the model computational graph based on their causal contribution to the model behavior to obtain the representative circuit. This procedure could be directly applied to the weights of the neural network. However, iteratively testing the causal impact of each model component requires high computational time. As already mentioned, attribution patching~\cite{syed-etal-2024-attribution} deals with this efficiency problem by computing the gradient of the performance measure we want to consider with respect to the model activations, limiting the accuracy in the identification of activations that are causally responsible for the model behavior.

\subsection*{Sensitivity}

The sensitivity of mechanistic techniques to implementation choices is an open challenge. The causal responsibility attributed to a specific component often depends on the data used during intervention. A component identified as responsible for a behavior under one set of inputs may exhibit negligible causal impact under different data~\cite{wang2024grokking}.
Circuit tracking approaches inherit this data sensitivity. Since circuit extraction relies on causal intervention, the resulting circuit may vary across input data. Furthermore, circuit discovery methods rely on thresholds to determine component inclusion in the circuit~\cite{conmy2023towards,syed-etal-2024-attribution}. Small perturbations to these thresholds can yield different structures. Addressing these limitations is an important research direction for the mechanistic interpretability of deep neural networks.

%%%%%%%%%%%%%%%%%%%%%%%%%%%%%%%%%%%%%%%%%%%%%%%%%%%%%%%%%%%%%%%%%%%%%%%%%%%%%%
% ---------------------------------------------------
% Acknowledgments
% ---------------------------------------------------
%
% \section*{Acknowledgment}

% ---------------------------------------------------
% References
% ---------------------------------------------------
%


\newpage

\section*{Biographies}
\label{sec:bio}

\begin{IEEEbiographynophoto}{Lorenzo Cirillo}
received the B.S. degree in computer and control engineering, the M.S. degree in artificial intelligence and robotics, earning the Excellent Student Award, from the Sapienza University of Rome, Italy, in 2022 and 2024, respectively, where he is currently pursuing the Ph.D. degree in cybersecurity at the AlcorLab. In 2023, he was an intern researcher at the Bank of Italy, working on knowledge graphs and graph databases. He is a member of the EURASIP Biometrics, Data Forensics, and Security Technical Area Committee. His research interests include deep learning, computer vision, adversarial machine learning, and explainable artificial intelligence.  
\end{IEEEbiographynophoto}

\begin{IEEEbiographynophoto}{Irene Amerini}
is an Associate Professor at DIAG, Sapienza University of Rome, where she leads the Computer Vision and Multimedia Forensics Research Team at ALCORLab. She was an Assistant Professor at Sapienza and a postdoctoral researcher at the Image Forensics and Security Lab, University of Florence. She received her Ph.D. from the University of Florence in 2011 and held visiting positions at Binghamton University and Charles Sturt University. She has served in IEEE, EURASIP, and IAPR committees, and is Senior Area Editor for IEEE Signal Processing Letters. Her research focuses on trustworthy AI, computer vision, adversarial learning, multimedia forensics.
\end{IEEEbiographynophoto}

\begin{IEEEbiographynophoto}{Andrea Cavallaro}
is a Full Professor at EPFL and a Fellow of the IAPR, ELLIS, and the Higher Education Academy. His research focuses on machine learning for multimodal perception, computer vision, machine listening, and information privacy. 
He served as an IEEE Signal Processing Society Distinguished Lecturer and as Chair of the IEEE Image, Video and Multidimensional Signal Processing Technical Committee. Furthermore, he has served as an Area, Associate, and Guest Editor for numerous leading journals and magazines, including IEEE Transactions on Image Processing, Signal Processing, and Multimedia, and the IEEE Signal Processing Magazine.
\end{IEEEbiographynophoto}

\vfill

\end{document}